\pdfoutput=1
\documentclass[10pt,twocolumn,letterpaper]{article}

\usepackage{cvpr}              

\usepackage{graphicx}
\usepackage{amsmath}
\usepackage{amssymb}
\usepackage{booktabs}
\usepackage{adjustbox}
\usepackage{makecell}
\usepackage{multirow}
\usepackage{bbding}

\usepackage[pagebackref,breaklinks,colorlinks]{hyperref}

\usepackage[capitalize]{cleveref}
\crefname{section}{Sec.}{Secs.}
\Crefname{section}{Section}{Sections}
\Crefname{table}{Table}{Tables}
\crefname{table}{Tab.}{Tabs.}

\newcommand\blfootnote[1]{%
  \begingroup
  \renewcommand\thefootnote{}\footnote{#1}%
  \addtocounter{footnote}{-1}%
  \endgroup
}

\def\cvprPaperID{*****} 
\def\confName{CVPR}
\def\confYear{2022}

\begin{document}

\title{Tip-Adapter: Training-free CLIP-Adapter for Better Vision-Language Modeling}


\author{Renrui Zhang$^{*1}$, Rongyao Fang$^{*2}$,  Wei Zhang$^{*1}$, Peng Gao$^{\dagger 1}$, Kunchang Li$^1$\\ Jifeng Dai$^3$, Yu Qiao$^1$, Hongsheng Li$^{2}$\\ 
  $^1$Shanghai AI Laboratory \quad \\
  $^2$The Chinese University of Hong Kong  \quad
  $^3$SenseTime Research\\
\texttt{\{zhangrenrui, gaopeng, qiaoyu\}@pjlab.org.cn} \\
\texttt{rongyaofang@link.cuhk.edu.hk, hsli@ee.cuhk.edu.hk}
}

\maketitle
\blfootnote{$^*$ indicates equal contributions, and $\dagger$ indicates corresponding author.}
\begin{abstract}
   Contrastive Vision-Language Pre-training, known as CLIP, has provided a new paradigm for learning visual representations by using large-scale contrastive image-text pairs. It shows impressive performance on zero-shot knowledge transfer to downstream tasks. To further enhance CLIP's few-shot capability, CLIP-Adapter proposed to fine-tune a lightweight residual feature adapter and significantly improves the performance for few-shot classification. However, such a process still needs extra training and computational resources. In this paper, we propose \textbf{T}raining-Free CL\textbf{IP}-\textbf{Adapter} (\textbf{Tip-Adapter}), which not only inherits CLIP's training-free advantage but also performs comparably or even better than CLIP-Adapter. Tip-Adapter does not require any back propagation for training the adapter, but creates the weights by a key-value cache model constructed from the few-shot training set.
   In this non-parametric manner, Tip-Adapter acquires well-performed adapter weights without any training, which is both efficient and effective. Moreover, the performance of Tip-Adapter can be further boosted by fine-tuning such properly initialized adapter for only a few epochs with super-fast convergence speed. We conduct extensive experiments of few-shot classification on ImageNet and other 10 datasets to demonstrate the superiority of proposed Tip-Adapter. 
   The code will be released at \url{https://github.com/gaopengcuhk/Tip-Adapter}.
   
\end{abstract}

\begin{figure}[t]
  \centering
\includegraphics[width=0.48\textwidth]{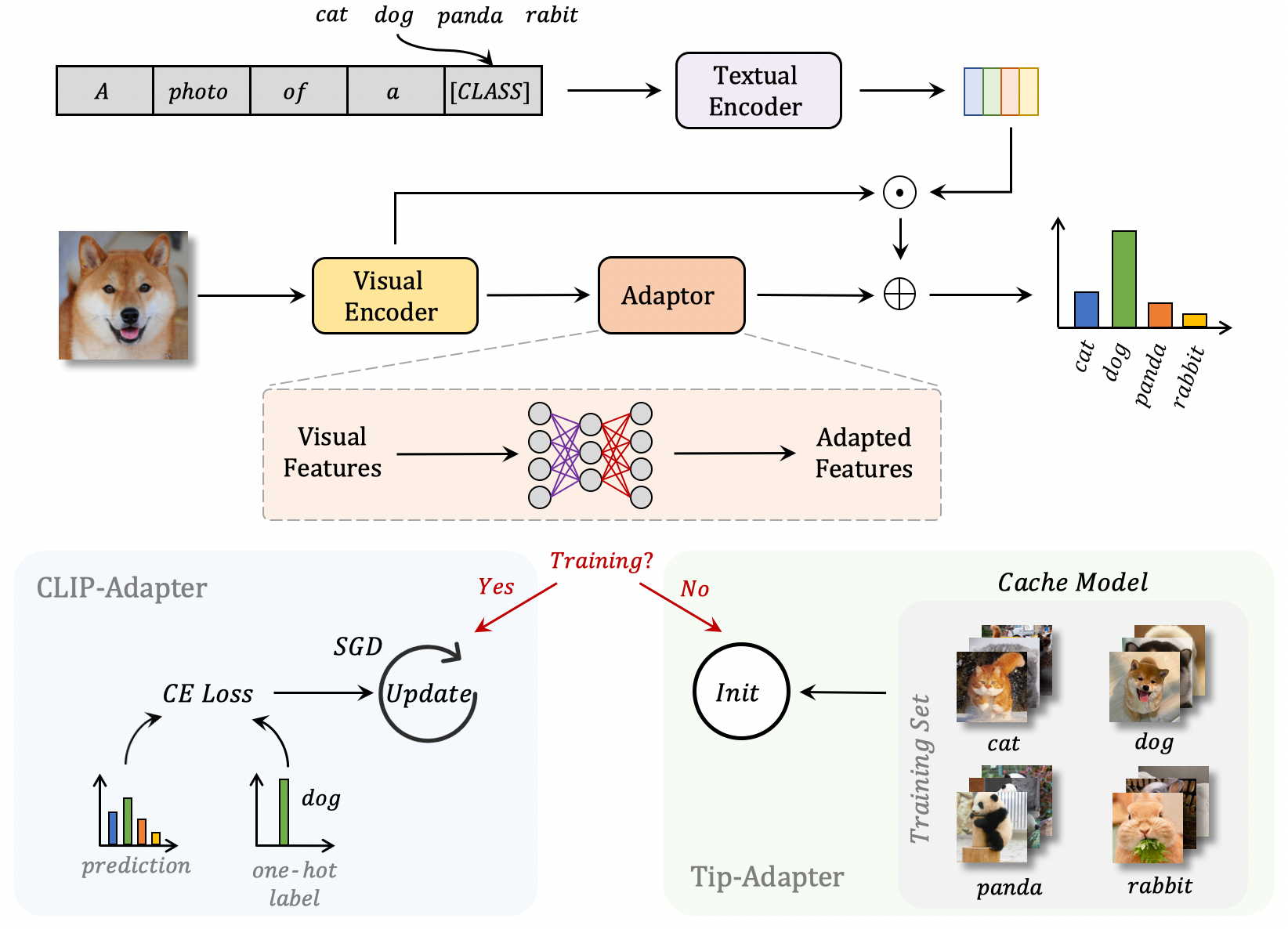}
   \caption{\textbf{A Comparison of CLIP-Adapter vs. the proposed Training-free CLIP-Adapter (Tip-Adapter).} CLIP-Adapter and Tip-Adapter share similar architectures, namely, two linear layers and a residual connection blending updated visual features with pre-trained CLIP features. CLIP-Adapter is trained with Stochastic Gradient Descent (SGD), while Tip-Adapter is training-free, whose weights of linear layers are initialized from Cache Model.}
    \label{fig:simple_intro}
    \vspace{-0.5cm}
\end{figure}

\section{Introduction}
\label{sec:intro}
Vision and language are two modalities for humans to perceive the surrounding world and perform diverse interactions with the environment. 
The accuracy of vision tasks, such as classification~\cite{krizhevsky2012imagenet, he2016deep,howard2017mobilenets,dosovitskiy2021vit,touvron2021training,mao2021dual}, detection~\cite{ren2015faster,carion2020end,gao2021fast} and segmentation~\cite{zheng2021rethinking}, has been boosted significantly thanks to better neural architecture designs (e.g. ResNet~\cite{he2016deep}, Transformer~\cite{vaswani2017attention}, etc.) and delicately designed frameworks (e.g., Faster R-CNN~\cite{ren2015faster}, RetinaNet~\cite{lin2017focal}, DETR~\cite{carion2020end}, etc.). 
Language generation and understanding have also been improved a lot due to large-scale self-supervised tasks, including mask prediction~\cite{devlin2018bert} and language model pre-training on web-scale dataset ~\cite{radford2018improving}. 
As vision and language usually contain complementary information, joint learning of vision and language representations has been proven to be quite effective on multi-modality tasks, such as Visual Question Answering ~\cite{antol2015vqa,anderson2018bottom,kim2018bilinear, gao2019dynamic}, Image Captioning~\cite{you2016image,huang2019attention}, and Referring Expression~\cite{yu2018mattnet}. 
Different from previous multi-modality interaction methods that learn vision and language representations independently on each dataset~\cite{anderson2018bottom,lu2019vilbert,tan2019lxmert}, CLIP~\cite{radford2021learning} proposed to learn transferable visual features from natural language supervisions and successfully demonstrate the amazing zero-shot classification ability of pre-trained model. Due to the interplay between language and vision, the encoded visual representations can be used in open-vocabulary classification without re-training. Following the research direction of prompt design~\cite{brown2020language,liu2021pre}, CoOp~\cite{zhou2021coop} proposed to fine-tune a pre-trained CLIP model for few-shot classification via learnable continuous tokens and achieved strong performance on few-shot image classification. Recently, 
CLIP-Adapter~\cite{gao2021clip} was also proposed to tackle the few-shot classification problem by introducing a feature adapter that generates the adapted features and then combines it with the original CLIP feature using a residual connection, which incorporates the new knowledge from the few-shot training set and prior knowledge encoded in CLIP. CLIP-Adapter demonstrates promising performance on few-shot classification without utilizing prompt designs.

Although CoOp~\cite{zhou2021coop} and CLIP-Adapter~\cite{gao2021clip} show strong performance on few-shot classification benchmarks, in comparison with CLIP~\cite{radford2021learning} and linear probe CLIP~\cite{radford2021learning}, they generally require much computational resources to fine-tune the large-scale vision-language model due to the slow convergence of Stochastic Gradient Descent (SGD)~\cite{kingma2014adam,loshchilov2017decoupled} and huge GPU memory consumption~\cite{rumelhart1985learning}.
Thus, we ask the following question: {Can we achieve best of both worlds of CLIP and CLIP-Adapter, which does not only has the advantage of training-free property of CLIP but also enjoys the strong performance of CLIP-Adapter
for few-shot classification?}

To achieve the goal, we propose a novel \textbf{T}raining-free CL\textbf{IP}-\textbf{Adapter} (\textbf{Tip-Adapter}), which adopts the architecture design of CLIP-Adapter. It appends CLIP model with an adapter of two-layer Multi-layer Perceptron (MLP) and a residual connection~\cite{he2016deep} combining pre-trained features with the updated features. Different from CLIP-Adapter, Tip-Adapter does not require SGD to train the adapter but constructs a query-key cache model~\cite{khandelwal2019generalization,orhan2018simple,grave2017unbounded} from few-shot supervisions to obtain the weights of adapter. Specifically, Tip-Adapter extracts visual features of few-shot training images by CLIP's visual encoder and transforms their corresponding labels into one-hot encoding. On top of that, a cache model is created which contains visual features and one-hot labels of the few-shot training set as key-value pairs.

Based on the constructed cache model, the weights of Tip-Adapter could be obtained in a training-free non-parametric manner. In detail, the two linear layers of the CLIP-Adapter are set as the cached visual features and their corresponding one-hot labels from the few-shot training set. Thus, the processing of the adapter with such weights can be regarded as retrieving the few-shot knowledge from the key-value cache model. By this approach, Tip-Adapter exhibits great efficiency compared to CLIP-Adapter's SGD training.
During inference, the test image's adapted feature is combined with its original CLIP-encoded feature. In this way, the Tip-Adapter is able to exploit knowledge from both the pre-trained CLIP and the few-shot training dataset.
Surprisingly, Tip-Adapter with such constructed weights could perform comparably to the fully fine-tuned CLIP-Adapter. 
Furthermore, if we unfreeze the first linear layer of the Tip-Adapter and further fine-tune it, Tip-Adapter's performance could be significantly boosted with just a few training epochs. It only requires 20 epochs in comparison with CLIP-Adapter's 200 epochs. 

The contributions of our paper are summarised below:
\begin{itemize}
    \item We propose Training-free CLIP-Adapter (Tip-Adapter), which has strong performance on few- classification via directly setting the weights of adapter with a cache model to avoid the conventional SGD fine-tuning. 
    
    \item The performance of Tip-Adapter can be further improved by fine-tuning based on such constructed weights with super-fast convergence.
    
    \item We evaluate Tip-Adapter on 11 few-shot classification datasets and conduct extensive ablation studies to demonstrate its characteristics. Tip-Adapter achieves competitive performance with state-of-the-art methods and leads to significant reduction of training time. 
\end{itemize}

\section{Related Work}
\label{sec:related work}

\paragraph{Data-efficient Transfer Learning.}
The capability of deep neural networks is revealed with the assistance of large-scale datasets~\cite{krizhevsky2012imagenet}. However, collecting large-scale dataset is challenging due to actual data's long-tail distribution, noisy annotations and the increasing labeling costs. Thus, transfer learning is proposed to reduce such costly requirements and has become a popular research field. Supervised pre-training on large-scale image classification tasks (e.g. ImageNet~\cite{deng2009imagenet}) and transferring to downstream tasks (e.g. detection~\cite{ren2015faster} and segmentation~\cite{he2017mask}) have been widely adopted.
Self-supervised learning, such as MoCo~\cite{he2020momentum} and BYOL~\cite{grill2020bootstrap}, further alleviates the need of large-scale supervised pre-training and converts the problem into a self-supervised form. From the view of language, VirTex~\cite{desai2021virtex} verifies the data-efficiency of language supervision via captioning on learning high-quality visual representations. Recently, CLIP~\cite{radford2021learning}, DeCLIP~\cite{li2021supervision} and ALIGN~\cite{jia2021scaling} have demonstrated that simple contrastive learning between vision-language pairs is able to learn zero-shot transferable features over diverse image classification datasets without further training. Furthermore, CoOp~\cite{zhou2021coop}, CLIP-Adapter~\cite{gao2021clip} and WiSE-FT~\cite{wortsman2021robust} indicate that the performance of CLIP can be significantly improved with either abundant or limited training data by fine-tuning weight-fixed CLIP with adapter or prompt optimization. In contrast, our proposed Tip-Adapter aims at directly infusing few-shot supervisions into the pre-trained CLIP model with a training-free manner.
Thus, Tip-Adapter is quite efficient at time and computation, as it only calculates and caches the few-shot training set once, and requires no more training.

\begin{figure*}[ht!]
  \centering
    \includegraphics[width=1\textwidth]{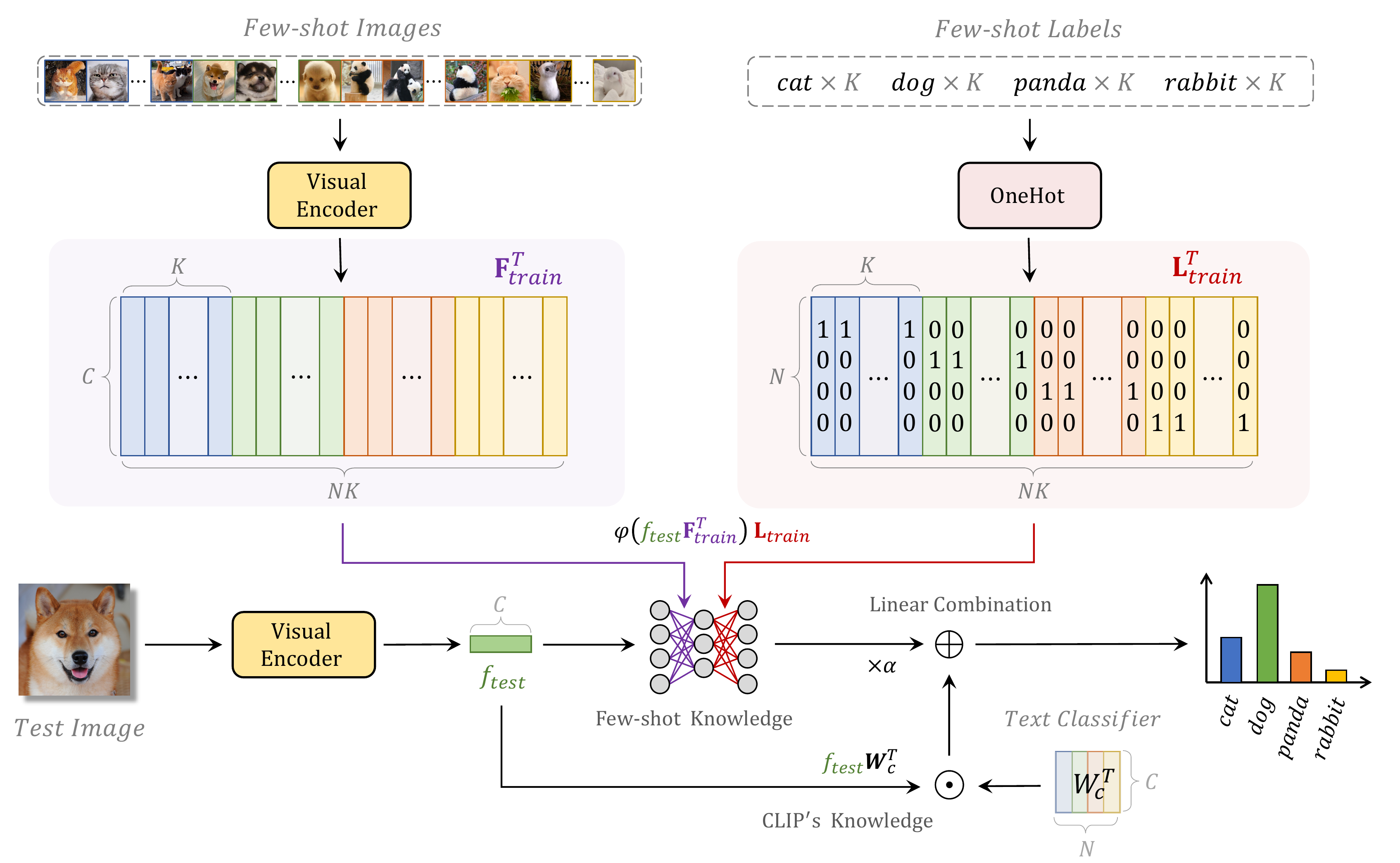}
   \caption{\textbf{The Pipeline of Tip-Adapter.} Given a $K$-shot $N$-class training set, we construct the weights of the two-layer adapter by creating a cache model from the few-shot training set. It contains few-shot visual features $F_{train}$ encoded by CLIP's visual encoder and few-shot ground-truth labels $L_{train}$. $F_{train}$ and $L_{train}$ can be used as the weights for the first and second layers in the adapter.}
    \label{fig:cache_model}
    \vspace{-0.5cm}
\end{figure*}
\paragraph{Transformer.}

Transformer~\cite{vaswani2017attention} introduces the key-query attention mechanism for encoding high-quality features and is superior to many previous language and vision models, such as Long Short-term Memory (LSTM)~\cite{hochreiter1997long} and Convolution Neural Network (CNN)~\cite{lecun1998gradient}. For natural language processing, transformer achieves leading performance on machine translation~\cite{vaswani2017attention,ott2018scaling}, natural language understanding~\cite{devlin2018bert} and language generation~\cite{radford2018improving,brown2020language}. Also for vision, transformer-based architectures have shown great capabilities and potentials on image classification~\cite{dosovitskiy2020image}, object detection~\cite{carion2020end}, and semantic segmentation~\cite{zheng2021rethinking}. Transformer utilizes multi-head key-query multiplication to implement highly-efficient message passing between all pairs of input tokens. In this paper, we analyse the relationship between Tip-Adapter's cache model and transformer's attention mechanism. 
This similar relations have also been discussed in persistent memory~\cite{sukhbaatar2019augmenting} and non-parametric attention~\cite{kossen2021self}.

\paragraph{Cache Model.}
A cache model stores features of training images and their labels as a key-value database. During inference, the feature generated from a test example would be treated as query and aggregate information from the database by retrieval ~\cite{vaswani2017attention}, which is similar to attention mechanism in transformer~\cite{vaswani2017attention}. It is important that the whole process is non-parametric~\cite{kossen2021self} and requires no parameter update. The cache model has been adopted for improving language generation in kNN-LMs~\cite{khandelwal2019generalization}, Unbounded Cache~\cite{grave2017unbounded} and Pointer Sentinel Mixture Models~\cite{merity2016pointer}. Although simple cache model~\cite{orhan2018simple} has shown some promising results, it needs to store a huge memory of training dataset for test-time inference. Therefore, approximate kNN and high-optimized similarity search system~\cite{johnson2019billion} are proposed, but the whole pipeline is generally verbose, error-prone and slow. Different from previous cache models' setup, we explore it with CLIP and adopt the few-shot settings. In detail, we generate cached features from CLIP's visual encoder so that strong label-related information is naturally encoded in the key-value pair. Due to the few-shot setup with limited training samples, the total cache is small and can be efficiently calculated by two cascades of matrix multiplication. Moreover, the cache model in Tip-Adapter is able to dynamically update via Stochastic Gradient Descent (SGD), which further improves its performance.

\section{Method}
\label{sec:method}
In this section, we introduce the proposed Training-free CLIP-Adapter (Tip-Adapter) for significantly improving CLIP's zero-shot classification accuracy. In Section~\ref{adapter}, we first briefly revisit CLIP-Adapter. In Section~\ref{tip}, we present the designs of Tip-Adapter and the further fine-tuned version for achieving stronger performance. Finally in Section~\ref{cache}, we discuss the relationship with previous cache-based methods.

\subsection{A Revisit of CLIP-Adapter}
\label{adapter}

Adapter \cite{houlsby2019parameter} is a lightweight neural module to conduct parameter-efficient fine-tuning of pre-trained models, such as BERT \cite{devlin2018bert} in natural language processing. 
CLIP learns visual representations by exploiting contrastive learning on large-scale image-text pairs, which achieves inspirational performance in zero-shot classification. Following the spirit of adapter, to transfer CLIP's knowledge for achieving few-shot classification, CLIP-Adapter~\cite{gao2021clip} appends a lightweight two-layer Multi-layer Perceptron (MLP) to the pre-trained fixed-weight CLIP model and predicts the adapted feature residuals for each input image. 
The residual connection in the adapter helps to fuse the prior visual knowledge encoded by CLIP and the updated features learned from the few-shot training set.
Compared with zero-shot and linear-probe CLIP ~\cite{radford2021learning}, CLIP-Adapter achieves significant performance improvement on multiple image classification datasets. It is optimized by minimizing the cross-entropy loss between predictions and ground truths with stochastic gradient descent (SGD). Thus, although CLIP-Adapter achieves great performance boost, it still requires additional training, which is slow and costly compared with the original training-free zero-shot CLIP.

Specifically, for an input image, its L2 normalized feature $f_c$ is first obtained by CLIP's pre-trained visual encoder. An adapter of two-layer MLP, with parameters $W_1, b_1, W_2, b_2$, is appended to obtain the updated feature $f_a$,
\begin{align}
\begin{split}
    f_a = \varphi(f_c W_1^T + b_1) W_2^T + b_2,
\end{split}
\end{align}
where $\varphi$ denotes the activation function in the MLP. Then, the adapted feature $f_a$ is linearly combined with the pre-trained feature $f_c$ with a hyper-parameter $\alpha \in [0,1]$ to output the final classification logits. In this way, the prior knowledge of CLIP on the input image is updated by the adapter in an additive manner,
\begin{align}
\label{v1}
\begin{split}
    \mathrm{logits} = \alpha f_a W_c^T + f_c W_c^T,
\end{split}
\end{align}
where $W_c$ is the weights of the text classifier. To construct $W_c$, following zero-shot CLIP, CLIP-Adapter places each category name into the pre-defined prompt template and encodes them by CLIP's pre-trained textual encoder.


\subsection{Training-free CLIP-Adapter}
\label{tip}

We propose Tip-Adapter, which is a training-free and non-parametric extension of CLIP-Adapter~\cite{gao2021clip} but performs comparably and even better than it. To achieve this goal, it adopts the same architecture as CLIP-Adapter, but we
construct a key-value cache model from the few-shot training set and transform the cache into the weights of the adapter MLP in a non-parametric manner without fine-tuning. 
Surprisingly, with weights constructed by a properly designed approach, Tip-Adapter without fine-tuning can achieve comparable performance as CLIP-Adapter with fine-tuning.
In addition, if fine-tuning is allowable, further fine-tuning with such weights as network initialization is able to achieve much higher accuracy with super-fast convergence speed.

\paragraph{Cache Model Construction.}
Given the pre-trained CLIP~\cite{radford2021learning} model and a $K$-shot $N$-class training set for few-shot classification, there are $K$ annotated images in each of the $N$ categories, denoted as $\mathrm{I}_{K}$ with their labels $\mathrm{L}_{N}$. The goal is to conduct image classification on the $N$ classes. 
We create a key-value cache model and convert it to obtain the weights of the proposed Tip-Adapter.
For each training image, we utilize the CLIP's visual encoder to extract its $C$-dimensional L2 normalized visual feature, and convert its ground-truth label into an $N$-dimensional one-hot vector. 
For all $NK$ training samples, we denote their visual features and corresponding label vectors as $\mathbf{F}_{\rm train} \in \mathbb{R}^{NK \times C}$ and $\mathbf{L}_{\rm train} \in \mathbb{R}^{NK \times N}$,
\begin{align}
    &\mathbf{F}_{\rm train} = \mathrm{Visual Encoder}(\mathrm{I}_
    {K}),\\
    &\mathbf{L}_{\rm train} = \mathrm{One Hot}(\mathrm{L}_{N}).
\end{align}
To create the key-value cache, the CLIP-encoded representations $\mathbf{F}_{\rm train}$ are treated as keys, while the one-hot ground-truth vectors $\mathbf{L}_{\rm train}$ are used as their values.
In this way, the key-value cache contains all the new knowledge extracted from the few-shot training set, which can be converted to the weights of the adapter MLP to update the prior knowledge encoded in the pre-trained CLIP.

\paragraph{Tip-Adapter.}
After constructing the cache model, the adapter weights are determined according to cached key-value pairs.
Specifically, the test image's normalized visual feature $f_\mathrm{test} \in \mathbb{R}^{1 \times C}$ is first extracted by the CLIP's visual encoder and serves as a query for retrieving from the key-value cache. The affinities between the test query and keys of the cached few-shot training set can be then estimated as $A \in \mathbb{R}^{1 \times NK}$,
\begin{align}
\label{beta}
    A 
       =  \exp{(-\beta(1 - f_\mathrm{test} \mathbf{F}^T_{\rm train}))},
\end{align}
where $\beta$ stands for a modulating hyper-parameter. Since both query and key features are L2 normalized, the term $(1-f_\mathrm{test} \mathbf{F}^T_{\rm train})$ is equivalent to calculating the Euclidean distances between the test feature $f_\mathrm{test}$ and all few-shot training images' features $\mathbf{F}^T_{\rm train}$. 
The exponential function is adopted to convert query-key Euclidean distances to non-negative affinities $A$ with $\beta$ modulating its sharpness. 
Afterwards, the retrieved value from the cache model can be obtained via the multiplication between the query-key affinities and the cached values as $A \mathbf{L}_{\rm train}$.


The predicted logits of the test image by the Tip-Adapter is then calculated as
\begin{align}
\label{logits}
    \mathrm{logits} &= \alpha A \mathbf{L}_\mathrm{train} + f_\mathrm{test} W_c^T \notag \\
    &= \alpha \varphi(f_\mathrm{test} \mathbf{F}^T_{\rm train}) \mathbf{L}_\mathrm{train} + f_\mathrm{test} W_c^T,
\end{align}
where $W_c$ represents CLIP's text classifier, $\alpha$ denotes the residual ratio, and we define $\varphi(x) = \exp(-\beta(1 - x))$.
Intuitively, Tip-Adapter's class prediction contains two terms: predictions according to values retrieved from the cache model and predictions from the pre-trained CLIP. The former term adaptively summarizes information from the few-shot training set. The values $\mathbf{L}_\mathrm{train}$ (class predictions of the cached samples) in the cache are linearly combined according to the query-key affinities $A$. The latter term preserves the prior knowledge from the original CLIP by directly using the pre-trained classifier $W_c^T$ to process the test image feature $f_\mathrm{test}$. The two terms are balanced by the weight $\alpha$. Empirically, $\alpha$ is set to be large if the domain gap between pre-trained and downstream tasks is large, since more knowledge from the few-shot set is required, and small otherwise.

Comparing Eqs.~\eqref{v1} and \eqref{logits}, the proposed Tip-Adapter can be seen as a special form of CLIP-Adapter with the following weights,
\begin{align}
\label{diffs}
    &W_1 = \mathbf{F}_\mathrm{train},  W_2 = \mathbf{L}^T_\mathrm{train},
   \ b_1 = 0, \ b_2 =0, \\
    &\varphi(x) = \operatorname{exp}(- \beta (1 - x)), \  \text{ where } \  x \in [0, 1].
\end{align}
The differences between CLIP-Adapter and Tip-Adapter can be summarized as follows. Firstly, CLIP-Adapter randomly initializes $W_1$ and $W_2$ and learns them via SGD, while Tip-Adapter directly sets $W_1$ as cached training features $\mathbf{F}_\mathrm{train}$ and $W_2$ as the transposed one-hot encoding of the ground-truth labels $\mathbf{L}_\mathrm{train}$, which are non-parametric and training-free. 
Secondly, the bottleneck dimension of Tip-Adapter is equal to $NK$, while CLIP-Adapter selects a low-dimensional bottleneck to prevent the risk of over-fitting. This indicates that with such proper initialization, the over-fitting problem on few-shot datasets is much alleviated, which further releases the fitting power of high-dimensional linear layers.
Thirdly, Tip-Adapter introduces a new activation function denoted in Eq.~\eqref{diffs}. As its inputs are the distances in the normalized feature space, it is bounded between 0 and 1. However, for CLIP-Adapter, the common activation function, ReLU$(\cdot)$, is chosen to handle unbounded inputs. Aided by the new activation $\varphi(\cdot)$, the calculated distances can be well modulated and help the performance.
In summary, Tip-Adapter could acquire the well-performed adapter weights without training. In other words, it is more efficient and effective on few-shot classification.




Tip-Adapter can greatly improve the classification performance of CLIP by incorporating new knowledge provided by the few-shot training set and be implemented as a CLIP-Adapter with cached $W_1$ and $W_2$. 
However, given more shots, there still exists slight performance gap between Tip-Adapter and CLIP-Adapter. 
To mitigate the gap, we can treat the cache model as a good initialization point and continue to fine-tune the Tip-Adapter via SGD to surpass the randomly initialized CLIP-Adapter.

For fine-tuning, we supervise the Tip-Adapter's predictions with few-shot training data and the cross-entropy loss, during which the weights in the cache model are updated via SGD. Specifically, we unfreeze the weights of keys $W_1 = \mathbf{F}_\mathrm{train}$, but still fix the weights of value $W_2 = \mathbf{L}_\mathrm{train}$ and two encoders of the CLIP model. The intuition is that updating the keys in the cache model can adaptively boost the estimation of affinities, namely, the distances calculation between training and testing images in the embedding space. In contrast, values in the cache are one-hot encodings representing ground-truth annotations and shall be kept frozen to accurately memorize the memory information. Thanks to the good initialization provided by the cache model, Tip-Adapter only needs to be fine-tuned for a small number of epochs and can surpass the CLIP-Adapter with much longer training scheme (20 epochs versus 200 epochs). It demonstrates that the proposed Tip-Adapter can achieve strong performance with fast convergence and limited resources.



\subsection{Relationship with Cache-based Networks}
\label{cache}

Constructing a cache model from few-shot training data has been explored by many previous methods, including Matching Network~\cite{vinyals2016matching}, Prototypical Networks~\cite{snell2017prototypical}, MAML~\cite{finn2017model}, Relation Network~\cite{sung2018learning} and others~\cite{dhillon2019baseline,chen2020new,tian2020rethinking,chen2019closer}. 
Different from all previous methods, Tip-Adapter adopts the residual architecture following CLIP-Adapter but proposes to use cached features for properly setting the weights of the CLIP-adapter. By performing further fine-tuning, Tip-Adapter significantly improves the performance of few-shot learning.

From another perspective, our Tip-Adapter can be regarded as a heterogenous cache model with both visual and textual features extracted by CLIP. Specifically, the weights $W_c$ of CLIP's linear classifier can be treated as cached textual features. Therefore, the final classification of the test image $f_\mathrm{test}$ is jointly inferred by multi-modality features, which fully exploit vision-language prior knowledge encoded in CLIP. From this perspective, the two terms in Eq.~\eqref{logits} can be reinterpreted as distance calculation with cached visual and textual features, respectively, and their importance is balanced by $\alpha$. 


\section{Experiments}
\label{sec:experiments}

\subsection{Training Settings}
We conduct experiments for Tip-Adapter on 11 image classification datasets: ImageNet \cite{deng2009imagenet}, StandfordCars \cite{krause20133d}, UCF101 \cite{soomro2012ucf101}, Caltech101 \cite{fei2004learning}, Flowers102 \cite{nilsback2008automated}, SUN397 \cite{xiao2010sun},  DTD \cite{cimpoi2014describing}, EuroSAT \cite{helber2019eurosat}, FGVCAircraft \cite{maji2013fine},  OxfordPets \cite{parkhi2012cats},  and Food101 \cite{bossard2014food}. Performance comparison is conducted between Zero-shot CLIP \cite{radford2021learning}, Linear-probe CLIP \cite{radford2021learning}, CLIP-Adapter \cite{gao2021clip} and CoOp \cite{zhou2021coop}.



As mentioned in the previous section, our Tip-Adapter has two versions. The first version is training-free, which directly sets the adapter's MLP weights following Eq.~\eqref{diffs}. The second version allows further fine-tuning of the adapter initialized by the properly set weights. The two version are denoted as \textit{Tip-Adapter} and \textit{Tip-Adapter-F} in this section, respectively.
Each model is trained with 1, 2, 4, 8, and 16 few-shot training sets, and tested on the full test sets. For the CLIP backbone, we utilize ResNet-50 \cite{he2016deep} as the visual encoder and a transformer \cite{dosovitskiy2021vit} as the textual encoder. In terms of prompt design, we adopt prompt ensembling in ~\cite{radford2021learning}, which inputs 7 templates into the CLIP textual encoder and then averages them as the final prompt. The 7 templates are: ``itap of a [CLASS].'', ``a bad photo of the [CLASS].'', ``a origami [CLASS].'', ``a photo of the large [CLASS].'', ``a [CLASS] in a video game.'', ``art of the [CLASS].'' and ``a photo of the small [CLASS].''.
To fine-tune Tip-Adapter-F, we train it with a batch size of $256$, and use Stochastic Gradient Descent (SGD)~\cite{kingma2014adam,loshchilov2017decoupled} with a learning rate $0.001$ and a cosine scheduler. In contrast to the 200-epoch training in CoOp and CLIP-Adapter, Tip-Adapter-F only requires 20 epochs for fine-tuning and has super-fast convergence speed, saving much computational cost for training. 

There are two image pre-processing methods adopted by existing methods. The first one is adopted by CLIP and the second one is reported in CoOp and CLIP-Adapter. We denote them as CLIP-style and CoOp-style pre-processing, respectively. Both of them are composed of random cropping, resizing, and random horizontal flip. They are different in resizing. The CLIP-style pre-processing resizes the cropped image's short side to 224 while keeping its original aspect ratio, while the CoOp-style resizes image's both sides to 224. We follow CLIP-style pre-processing by default, since it preserves cropped images' original aspect ratios.


\subsection{Comparison on ImageNet}

We compare our proposed Tip-Adapter and Tip-Adapter-F with {Zero-shot CLIP} \cite{radford2021learning}, {Linear-probe CLIP} \cite{radford2021learning}, {CoOp} \cite{zhou2021coop}, and {CLIP-Adapter} \cite{gao2021clip}. The zero-shot CLIP uses no extra training sample and conducts classification purely by calculating similarities between visual features of test images and textual features of the hand-crafted prompts. Linear-probe CLIP fine-tunes an additional linear classifier after the original frozen CLIP on the few-shot training set. To make prompts learnable, CoOp learns to generate different designs for the format of prompts, which we select its best-performance variant for comparison -- placing the class token at the end of the 16-token prompts and sharing the context among all classes. CLIP-Adapter appends a feature adapter \cite{houlsby2019parameter} on top of CLIP's visual encoder and textual encoder to narrow the domain gap between the pre-trained and transferred features, which helps to achieve better few-shot classification performance. Likewise, we compare Tip-Adapter with the best-performing variant of CLIP-Adapter with only learnable visual adapter. Moreover, for fair comparison, we adopt prompt ensembling of 7 templates for zero-shot CLIP and CLIP-Adapter, the same as Tip-Adapter.

\begin{figure}[t!]
\includegraphics[width=0.5\textwidth]{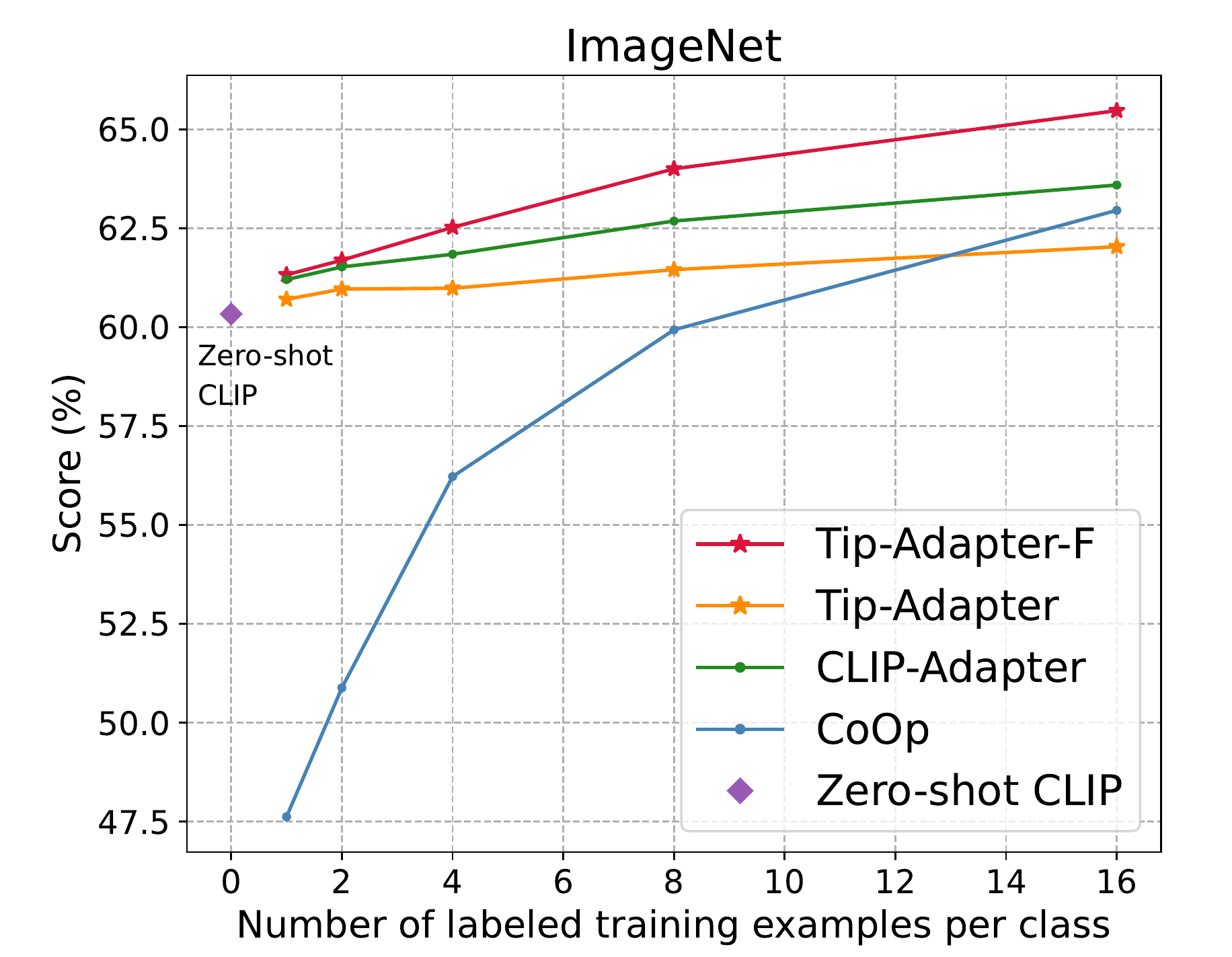}
   \caption{Classification accuracy of Tip-Adapter, Tip-Adapter-F and other models under different few-shot settings with CLIP-style pre-processing.}
    \label{imageneti}
    \vspace{-0.75cm}
\end{figure}

We report other models with both CLIP-style pre-process and CoOp-style pre-process in Table~\ref{imagenett} for fair comparison, but in Table~\ref{backbone}, Figure~\ref{imageneti} and the following pages, only the results with CLIP-style pre-processing are presented. Also, the far-behind performance of linear-probe CLIP are only presented in Figure~\ref{imageneti} for conciseness.

\paragraph{Performance Analysis.}
As shown in Figure~\ref{imageneti} and Table~\ref{imagenett}, both versions of our Tip-Adapter show outstanding performances over compared methods. Similarly without training, Tip-Adapter consistently surpasses zero-shot CLIP. Compared to linear-probe CLIP and CoOp with time-consuming fine-tuning, Tip-Adapter greatly exceeds them when the numbers of training samples are limited. For instance, in 1-shot and 2-shot settings, Tip-Adapter surpasses linear-probe CLIP by 38.53$\%$ and 29.06$\%$, and CoOp by 13.08$\%$ and 10.08$\%$. They demonstrate the superiority of our training-free cache-based method. However, Tip-Adapter still falls behind the well-tuned CLIP-Adapter. With further fine-tuning, Tip-Adapter-F, updates the weights initialized from the cache model and achieves the best performance over all methods in all few-shot settings. Compared to the training-free version, the performance gain resulted from fine-tuning becomes larger as the number of training samples increases, from 1-shot's +0.62$\%$ to 16-shot's +3.44$\%$. This accords with the intuition that more training samples provide the model with a more robust cache. 

\begin{table}[t!]
\centering
\vspace*{13pt}
\begin{adjustbox}{width=\linewidth}
	\begin{tabular}{lccccccc}
	\toprule
		Models & RN50 & RN101 &  ViT/32 & ViT/16 & RN50$\times$16\\ \midrule
		Zero-shot CLIP\cite{radford2021learning} &60.33 &62.53 &63.80 &68.73 &70.94\\
		CoOp\cite{zhou2021coop} & 62.95 &66.60 & 66.85  &71.92 &-\\
		CLIP-Adapter\cite{gao2021clip} & 63.59 & 65.39 & 66.19 &71.13 &-\\
		\textbf{Tip-Adapter} & 62.03 & 64.78 & 65.61 & 70.75 &72.95\\
	    \textbf{Tip-Adapter-F} & \textbf{65.51} & \textbf{68.56} & \textbf{68.65} & \textbf{73.69} &\textbf{75.81}\\
	\bottomrule
	\end{tabular}
\end{adjustbox}
\caption{Performances ($\%$) of different models on various vision backbones. RN50 denotes ResNet-50, and ViT/32 denotes ViT-Base with 32 $\times$ 32 patch size, and RN50$\times$16 denotes ResNet-50 with 16 times more computations~\cite{radford2021learning}. }
\vspace*{-12pt}
\label{backbone}
\end{table}

In Table~\ref{backbone}, we also implement Tip-Adapter with different visual encoders over ResNet~\cite{he2016deep} and Vision Transformer~\cite{dosovitskiy2021vit} backbones. The leading performances of Tip-Adapter-F fully demonstrate its superior cabability for visual recognition.


\paragraph{Efficiency Comparison.}
In Table~\ref{time}, we show the 16-shot training time of different models and the experiments are conducted on a single NVIDIA GeForce RTX 3090 GPU. Tip-Adapter and CLIP-Adapter first cache the textual features from the CLIP model and load them during training, but CoOp adopts learnable prompts so that the textual features are required to be calculated online in every iteration. Linear-probe CLIP utilizes Logistic Regression~\cite{wright1995logistic} to train the final linear layer, so it cannot be measured by epoch. From the comparison we can observe that CoOp takes most training time for learning prompts and has a 2.26$\%$ performance gain over zero-shot CLIP. CLIP-Adapter significantly reduces the training time with a larger improvement. Tip-Adapter gains 1.70$\%$ boost but requires no extra training time, which makes it a good trade-off between performance and efficiency. Tip-Adapter-F further reaches a higher accuracy with only $1/10$ of CLIP-Adapter's epochs, achieving best of both worlds.

\begin{table}[t]
\centering
\vspace*{13pt}
\begin{adjustbox}{width=\linewidth}
	\begin{tabular}{lcccccc}
	\toprule
		Few-shot Setup & 1    & 2  & 4  & 8 & 16 \\ \midrule
		\multicolumn{6}{c}{CoOp-style Pre-process,\quad Zero-shot CLIP$^{*}$\cite{radford2021learning}: \ 57.81} \\  \cmidrule(lr){1-6}
	   \cmidrule(lr){1-6}
		Linear-probe CLIP$^{*}$\cite{radford2021learning}  &21.14 &31.67 &40.33 &47.12 &51.59\\
		CoOp$^{*}$\cite{zhou2021coop}  &53.12 &53.23 &56.47 &57.58 &60.46 \\
	    CLIP-Adapter$^{*}$\cite{gao2021clip}  &58.17  &58.58  &59.40  &60.39 &61.33 \\ \midrule
	    \multicolumn{6}{c}{CLIP-style Pre-process,\quad Zero-shot CLIP\cite{radford2021learning}: \ 60.33} \\  \cmidrule(lr){1-6}
	    Linear-probe CLIP\cite{radford2021learning}  &22.17 &31.90 &41.20 &49.52 &56.13\\
		CoOp\cite{zhou2021coop}  &47.62 &50.88 &56.22 &59.93 &62.95 \\
	    CLIP-Adapter\cite{gao2021clip}  &61.20  & 61.52 &  61.84 &  62.68 &  63.59 \\
	    \textbf{Tip-Adapter}  &60.70   &60.96   &60.98   &61.45   &62.03 \\
	    \textbf{Tip-Adapter-F}  &\textbf{61.32}   &\textbf{61.69}   &\textbf{62.52} &\textbf{64.00}   &\textbf{65.51}  \\
	    \emph{by fine-tuning} &\color{blue}{+0.62} &\color{blue}{+0.73} &\color{blue}{+1.54} &\color{blue}{+2.55} &\color{blue}{+3.48} \\
	\bottomrule
	\end{tabular}
\end{adjustbox}
\caption{Classification accuracy ($\%$) of models under both CoOp-style and CLIP-style pre-processing in few-shot settings. Models following the CoOp-style pre-process method are labeled by $*$. The last row in blue records the performance gain brought by further fine-tuning over Tip-Adapter.}
\vspace*{-16pt}
\label{imagenett}
\end{table}

\begin{table}[t]
\centering
\vspace*{13pt}
\begin{adjustbox}{width=\linewidth}
	\begin{tabular}{lcccccc}
	\toprule
		Models & Epochs & Time &  Accuracy &Gain\\ \midrule
		Zero-shot CLIP\cite{radford2021learning} &0 &0 &60.33 &0\\
		Linear-probe CLIP\cite{radford2021learning} & - & 13min & 56.13 &\color{blue}{-4.20}\\
		CoOp\cite{zhou2021coop} & 200 &14h\ 40min & 62.95  &\color{blue}{+2.62}\\
		CLIP-Adapter\cite{gao2021clip} & 200 & 50min & 63.59 &\color{blue}{+3.26}\\
		\textbf{Tip-Adapter} & 0 & 0 & 62.03\ & \color{blue}{+1.70}\\
	    \textbf{Tip-Adapter-F} & 20 & 5min & 65.51 & \color{blue}{+5.18}\\
	\bottomrule
	\end{tabular}
\end{adjustbox}
\caption{Fine-tuning time and classification accuracy of 16-shot learning with different methods. The last column in blue records the performance gain relative to zero-shot CLIP.}
\vspace*{-12pt}
\label{time}
\end{table}

\begin{figure*}[ht!]
    \centering
    
    \begin{minipage}[t]{0.19\linewidth}
    \centering
    \includegraphics[width=1.3in]{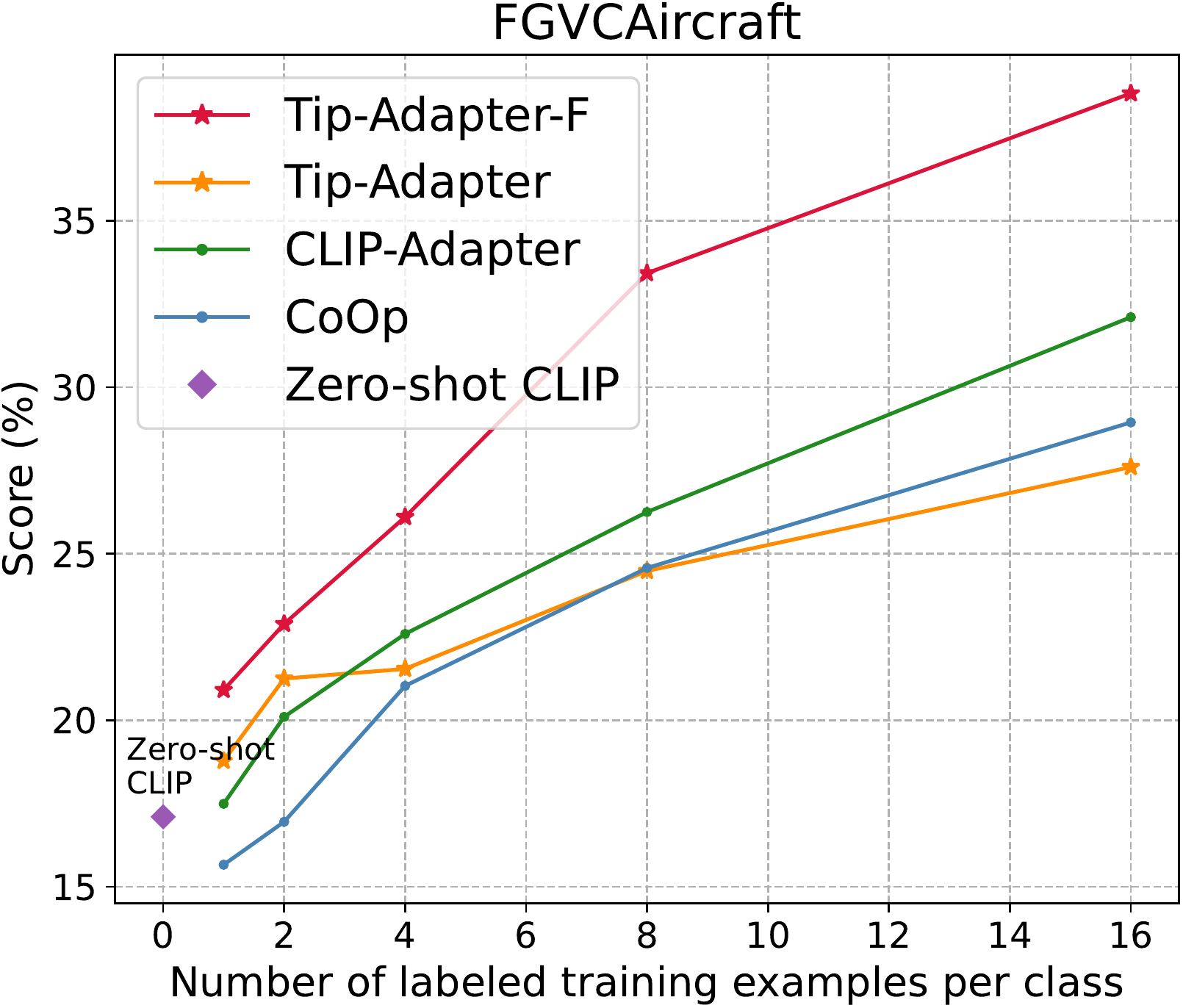}
    \end{minipage}
    \begin{minipage}[t]{0.19\linewidth}
    \centering
    \includegraphics[width=1.3in]{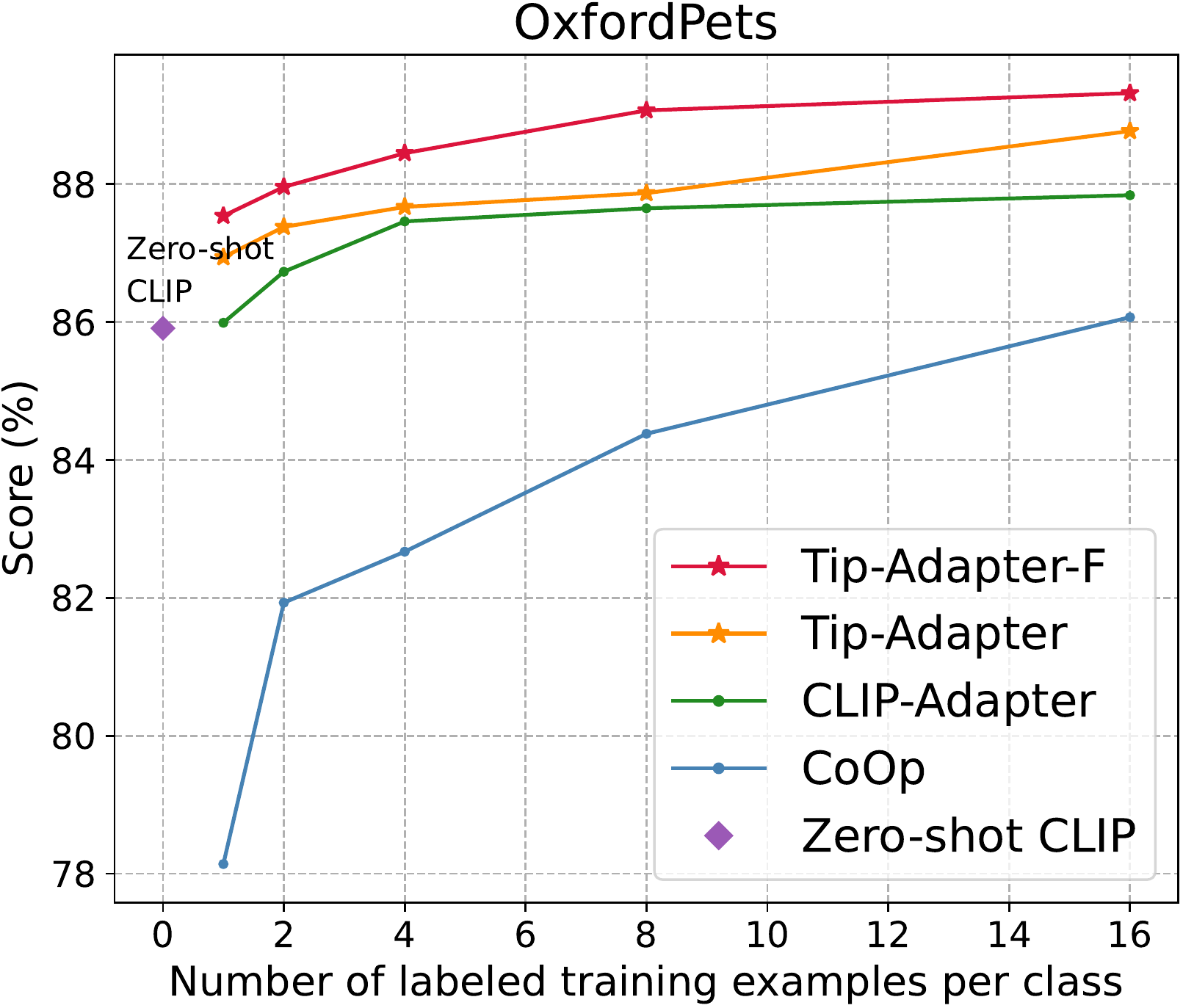}
    \end{minipage}
    \begin{minipage}[t]{0.19\linewidth}
    \centering
    \includegraphics[width=1.3in]{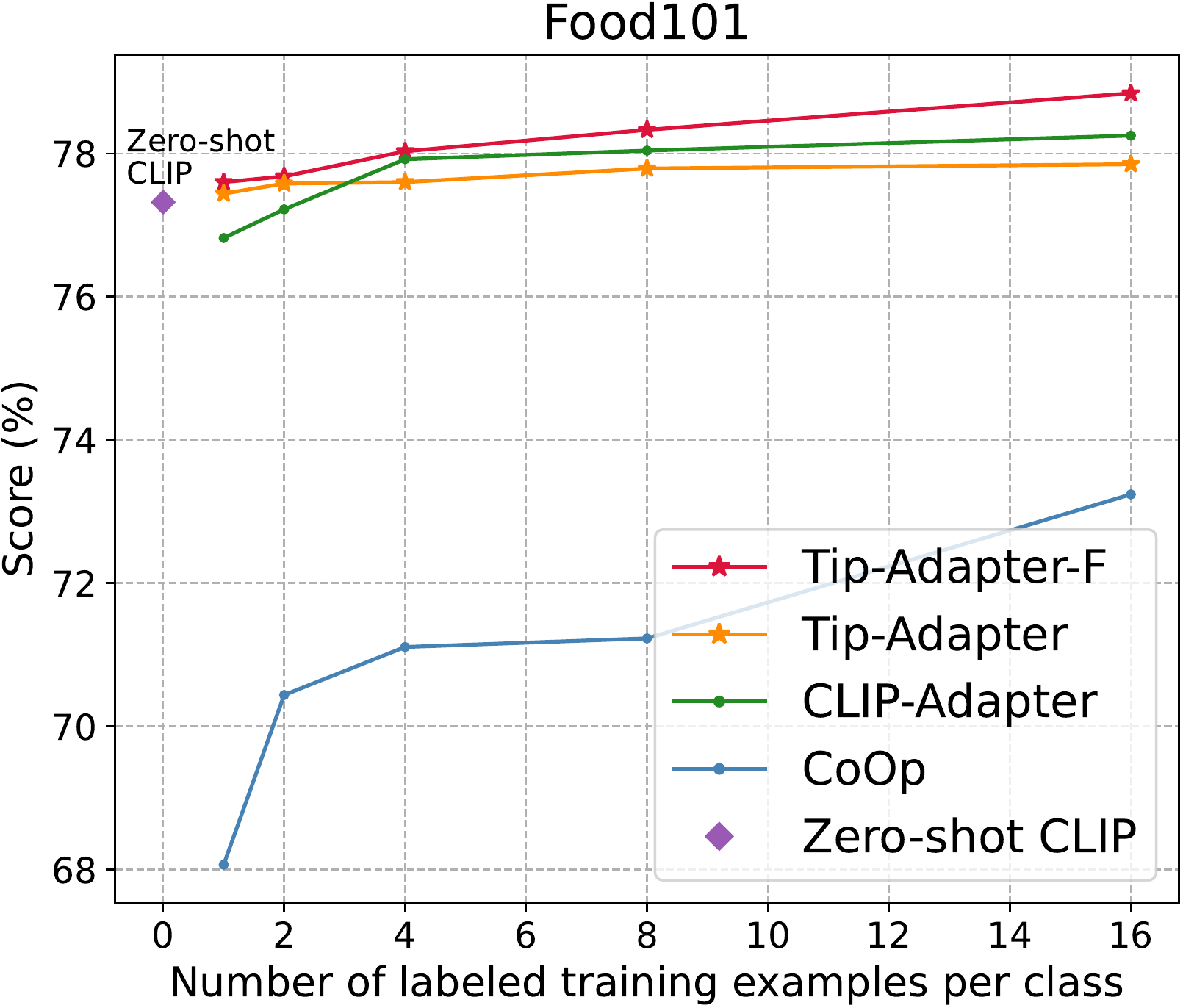}
    \end{minipage}
    \begin{minipage}[t]{0.19\linewidth}
    \centering
    \includegraphics[width=1.3in]{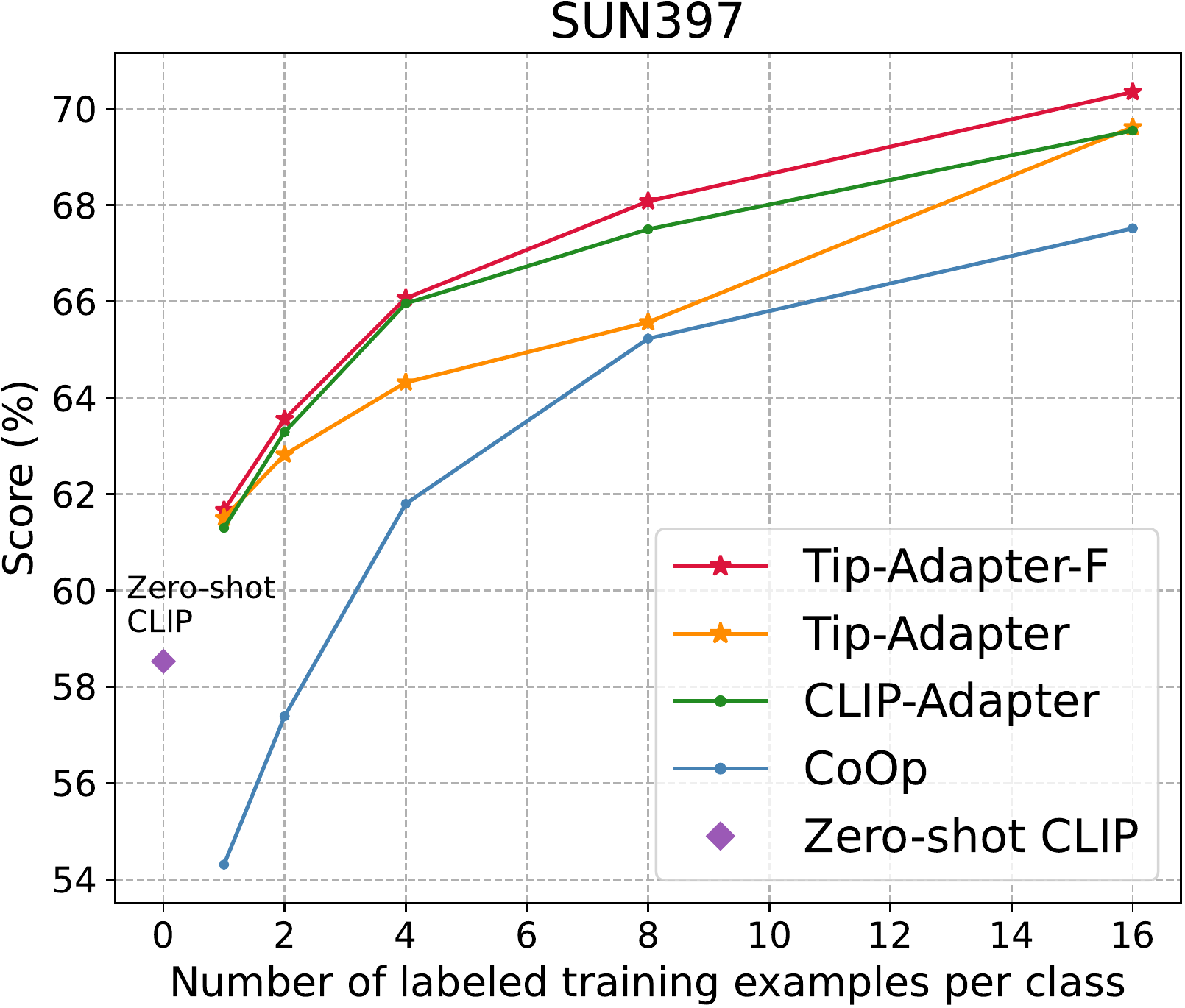}
    \end{minipage}
    \begin{minipage}[t]{0.19\linewidth}
    \centering
    \includegraphics[width=1.33in]{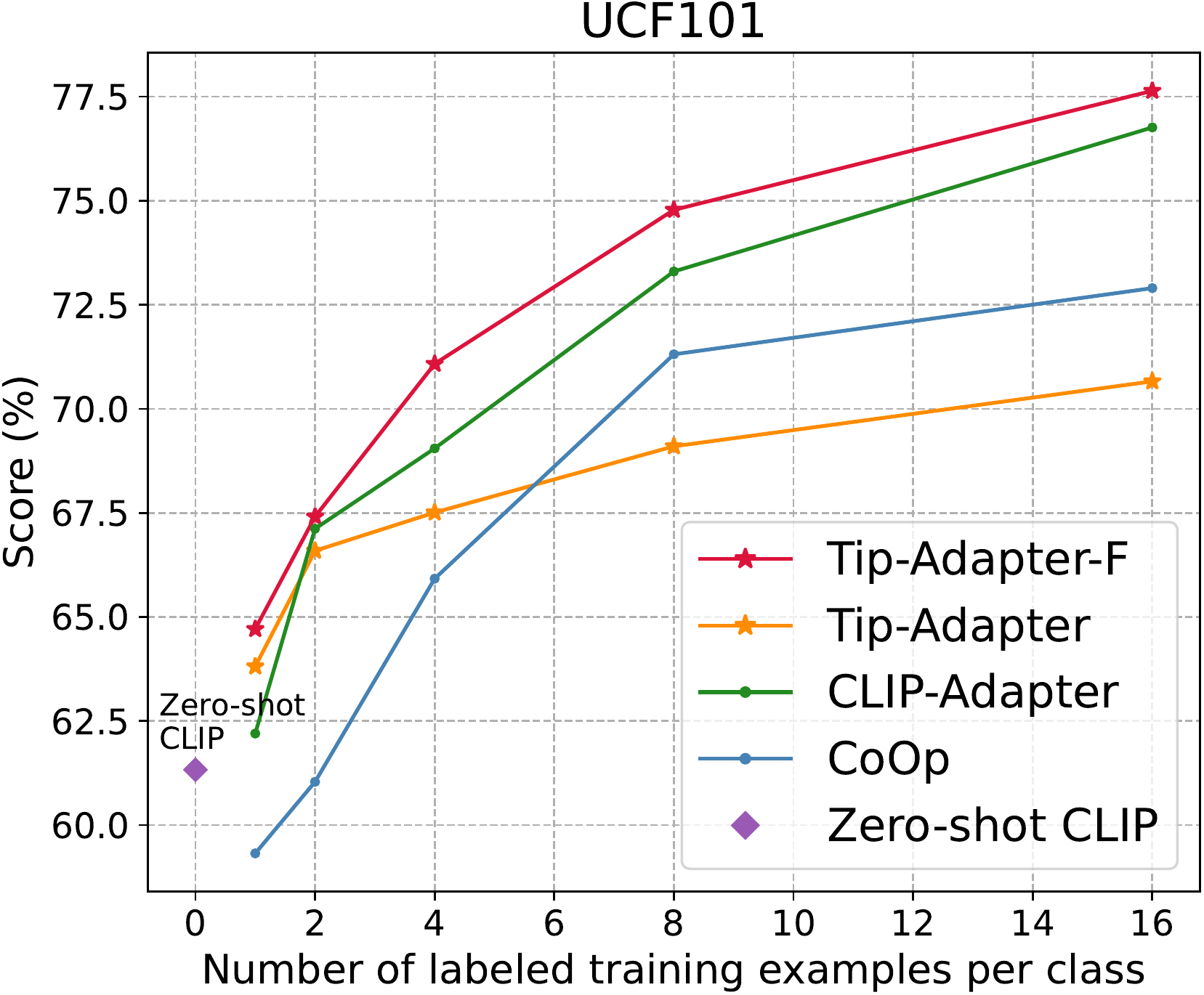}
    \end{minipage}

    \hspace{0.2in}
    
    \begin{minipage}[t]{0.19\linewidth}
    \centering
    \includegraphics[width=1.3in]{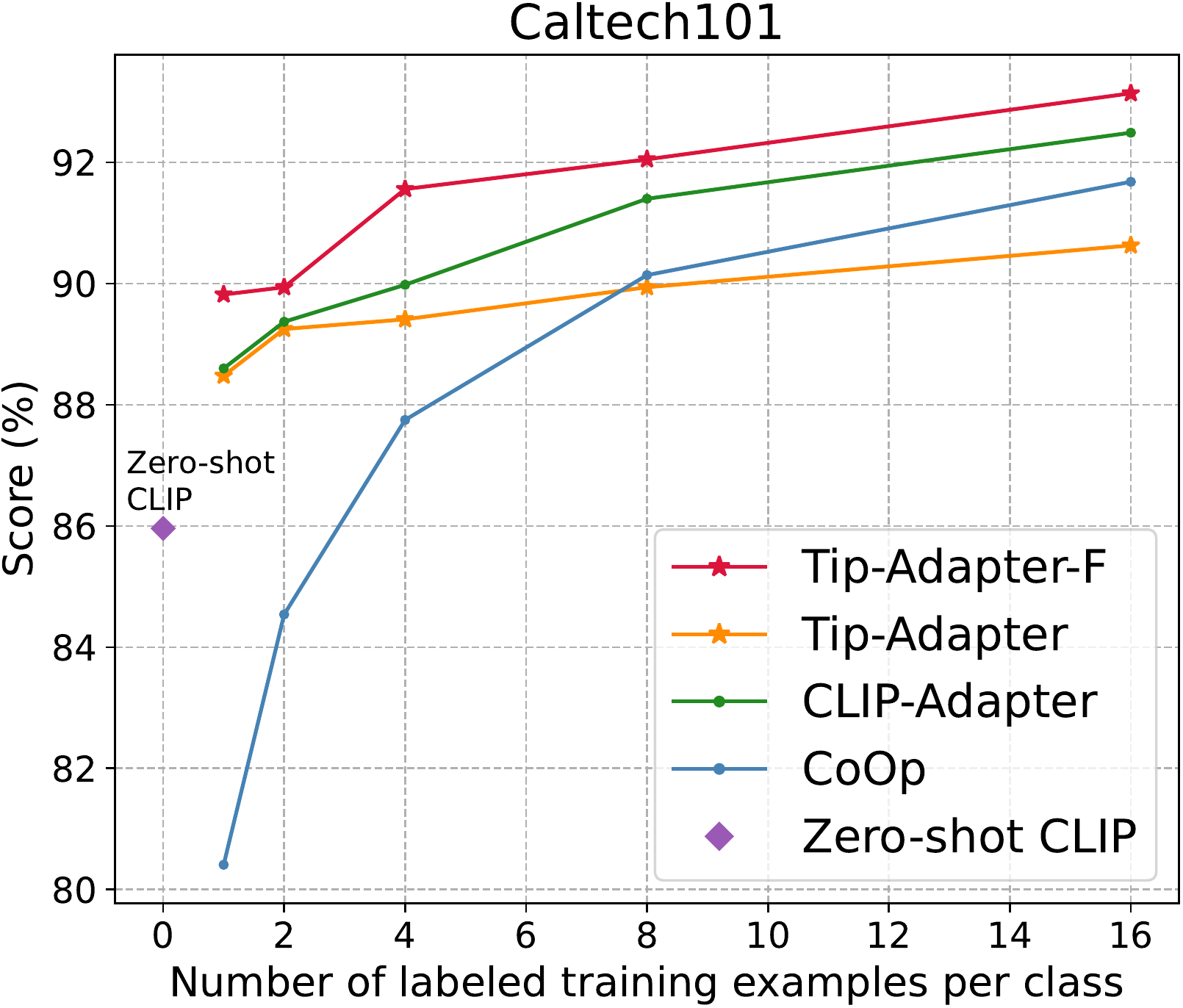}
    \end{minipage}
    \begin{minipage}[t]{0.19\linewidth}
    \centering
    \includegraphics[width=1.3in]{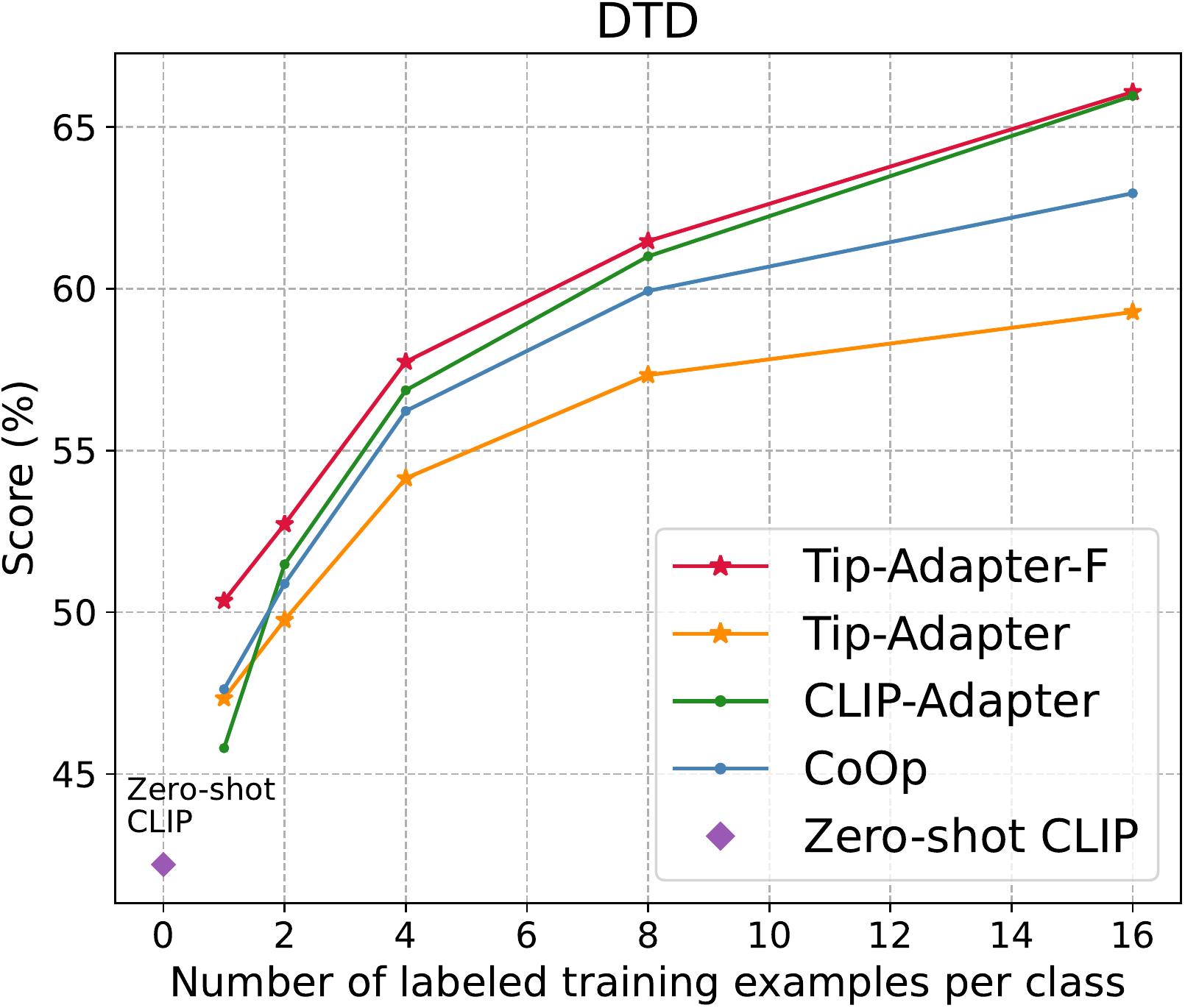}
    \end{minipage}
    \begin{minipage}[t]{0.19\linewidth}
    \centering
    \includegraphics[width=1.3in]{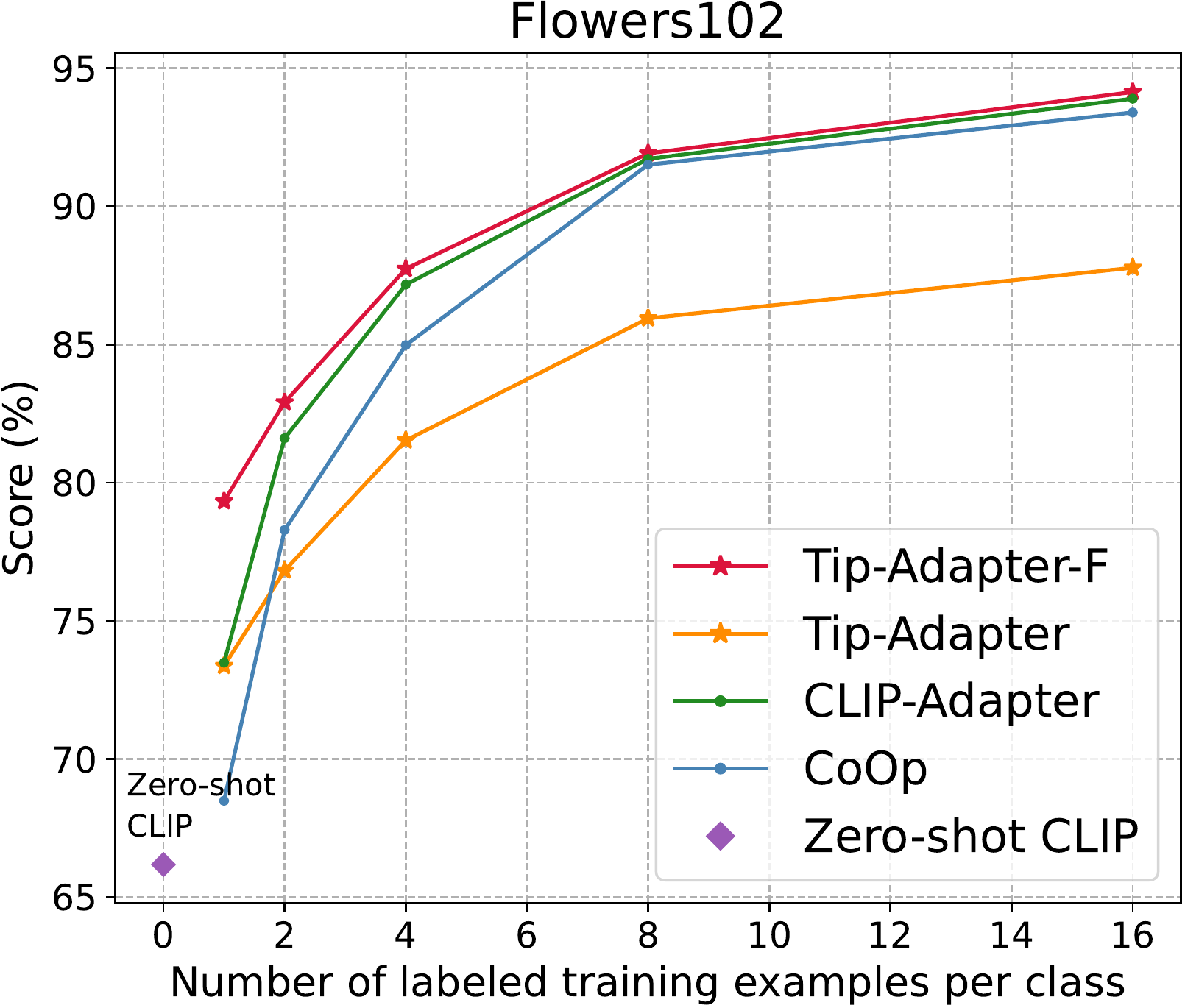}
    \end{minipage}
    \begin{minipage}[t]{0.19\linewidth}
    \centering
    \includegraphics[width=1.3in]{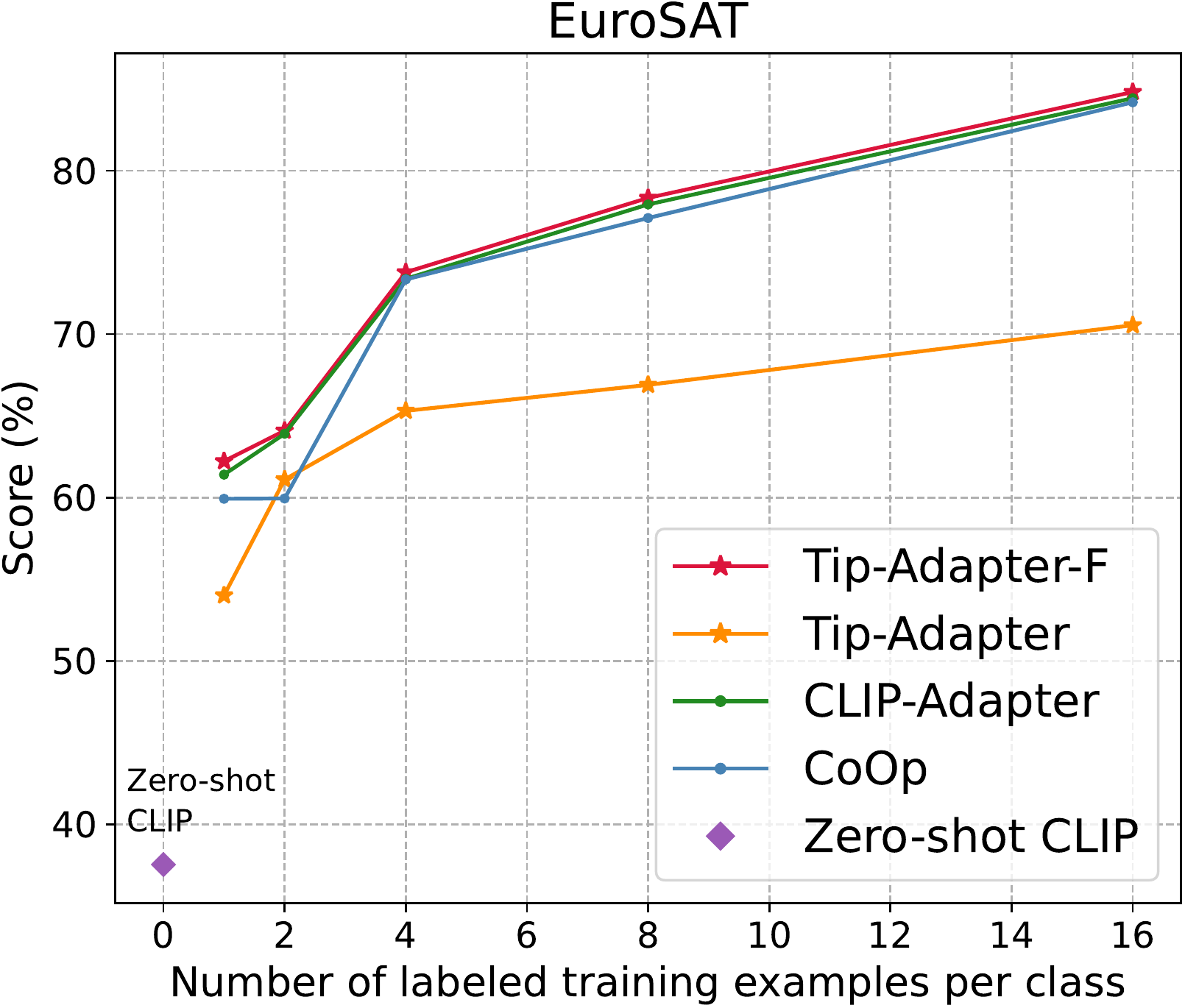}
    \end{minipage}
    \begin{minipage}[t]{0.19\linewidth}
    \centering
    \includegraphics[width=1.3in]{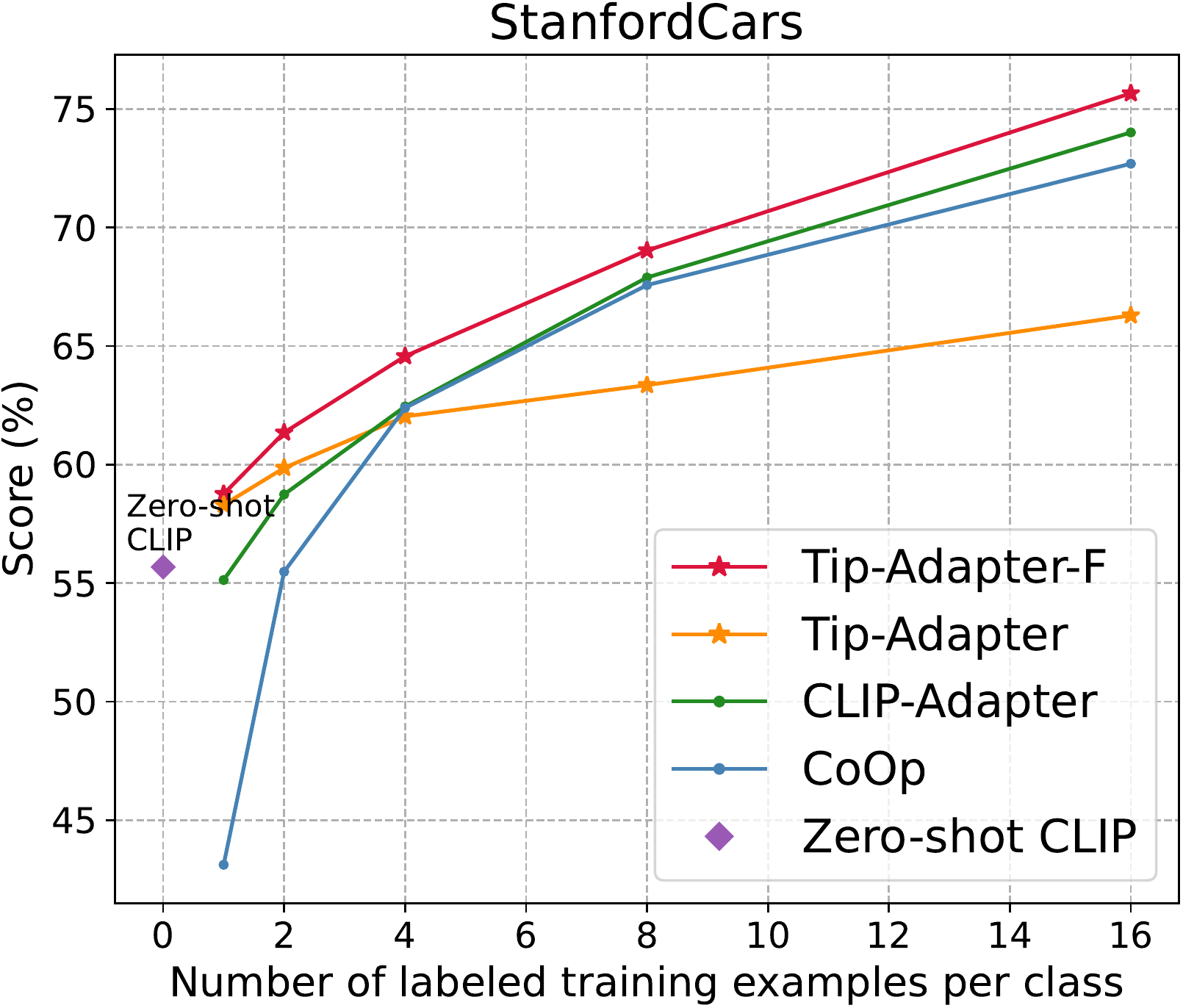}
    \end{minipage}
    
    \centering
    \caption{Results of few-shot classification on 10 datasets. Tip-Adapter exceeds zero-shot ClIP by a large margin, and Tip-Adapter-F further surpasses all compared methods by few-epoch fine-tuning.}
    \label{10 datasets}
    \vspace*{-6pt}
\end{figure*}

\subsection{Performances on Other Datasets}

Figure~\ref{10 datasets} shows Tip-Adapter's performances on other 10 datasets: StandfordCars \cite{krause20133d}, UCF101 \cite{soomro2012ucf101}, Caltech101 \cite{fei2004learning}, Flowers102 \cite{nilsback2008automated}, SUN397 \cite{xiao2010sun},  DTD \cite{cimpoi2014describing}, EuroSAT \cite{helber2019eurosat}, FGVCAircraft \cite{maji2013fine},  OxfordPets \cite{parkhi2012cats},  and Food101 \cite{bossard2014food}. Comparing with Zero-shot CLIP\cite{radford2021learning}, CLIP-Adapter\cite{gao2021clip} and CoOp\cite{zhou2021coop}, we observe that our Tip-Adapter significantly boosts the classification performance over zero-shot CLIP in different types of datasets. Although on some datasets, Tip-Adapter falls behind CLIP-Adapter when there are more shots for training, Tip-Adapter-F with a few-epoch fine-tuning eliminates this gap and further surpasses all other models. The consistent superiority over 10 datasets fully demonstrates the effectiveness and generality of Tip-Adapter.

\subsection{Ablation Studies}

In this section, we conduct several ablation studies about Tip-Adapter on ImageNet with the CLIP-style pre-processing. All experiments adopt the 16-shot setting and our training-free Tip-Adapter is adopted by default. 

\paragraph{Residual Ratio $\alpha$.}
The hyper-parameter $\alpha$ combines newly adapted features from cache model with pre-trained CLIP encoder's features, or in other words, weighing the importance of the visual and textual cache. As formulated above, larger $\alpha$ denotes using more knowledge from the few-shot training set and less otherwise. We vary $\alpha$ from 0.0 to 5.0, and set the hyper-parameter $\beta$ as 5.5. When $\alpha$ equals 0.0, the model is equivalent to zero-shot CLIP without using new knowledge from the training set. From the top part of Table~\ref{alpha}, we observe that the classification accuracy is improving as $\alpha$ increases from 0.0 to 1.0, achieving the best 62.03$\%$ at 1.0
, but further increasing the weight of the adapted features hurts the performance. This indicates that both the prior knowledge from CLIP and adapted features from the few-shot training set are equally important.


\begin{table}[t]
\centering
\vspace*{13pt}
\begin{adjustbox}{width=\linewidth}
	\begin{tabular}{c|cccccc}
	\toprule
		\multicolumn{7}{c}{Ablation Studies on Tip-Adapter} \\ 
		\midrule
		\multirow{2}{*}{Residual Ratio $\alpha$}  &0.0 &0.5 &\textbf{1.0} &2.0 &3.0 &4.0 \\  
        \cmidrule(lr){2-7}
		 &60.33  & 61.44  &\textbf{62.03} &61.41 &60.36 &59.14\\ 
        \midrule
		
	    \multirow{2}{*}{Sharpness Ratio $\beta$} &1.5 &3.5 &\textbf{5.5} &7.5 &9.5 &11.5 \\
	     \cmidrule(lr){2-7}
	    & 61.82  &61.91 &\textbf{62.03} &61.76 &61.62 &61.40\\
	    
	   \midrule
	   \multirow{2}{*}{Cache Size}&0 &1 &2 &4 &8 &\textbf{16} \\
	     \cmidrule(lr){2-7}
	    & 60.33  &61.45 &61.71 &61.79 &61.83 &\textbf{62.03}\\
	    
	    \midrule
	    \midrule
	   \multirow{3}{*}{\shortstack{\vspace*{2.2pt}\\More Shots\\\vspace*{0.3pt}\\than 16}} &\multicolumn{2}{l}{Shot Setup} &16 &32 &64 &128 \\
	     \cmidrule(lr){2-7}
	    &\multicolumn{2}{l}{Tip-Adapter} &62.03 &62.51 &62.88 &63.15\\
	     &\multicolumn{2}{l}{Tip-Adapter-F} &65.47 &66.58 &67.96 &69.74\\
	    
	\bottomrule
	\end{tabular}
\end{adjustbox}
\caption{Four ablation studies of the proposed Tip-Adapter on ImageNet, from top to bottom: residual ratio $\alpha$, sharpness ratio $\beta$, the size of cache model and the performance given more shots with cache size 16.}
\vspace*{-12pt}
\label{alpha}
\end{table}

\paragraph{Sharpness Ratio $\beta$.}
In Eq.~\eqref{beta}, $\beta$ in the activation function $\varphi$ controls the sharpness of the affinities. When $\beta$ is large, only the nearby training samples to the test image have large influences to its class prediction and vice versa.
The results of different $\beta$ are presented on the second part of Table~\ref{alpha}, in which $\alpha$ is set as 1.0. We observe that the variation of $\beta$ has limited impact. A moderate 5.5 for $\beta$ leads to the best-performing Tip-Adapter.


\begin{figure*}[ht!]
  \centering
    \includegraphics[width=0.95\textwidth]{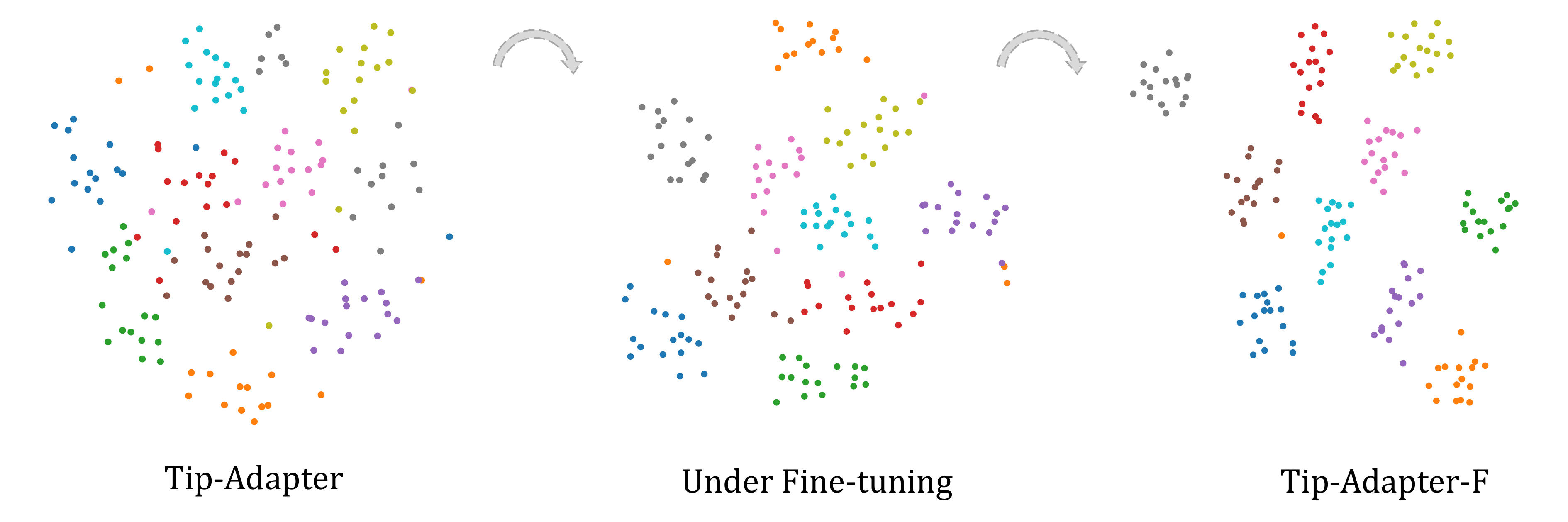}
   \caption{t-SNE Visualization of $W_1$ in Tip-Adapter. Dots in different colors stand for embeddings of different categories. From left to right, three distributions indicate the variation of $W_1$ during fine-tuning.}
    \label{tsne}
    \vspace{-0.5cm}
\end{figure*}

    

\paragraph{Size of the Cache Model.}
\label{size}
Here, we explore the influence of the cache model's size in Tip-Adapter. The largest setting caches 16 samples for each class, and the smallest one caches only one sample per class. We experiment with $K$ equaling 0, 1, 2, 4, 8 and 16. Cache size 0 is equivalent to zero-shot CLIP without the cache. When it is more than 0 but less than 16, taking 8 as an example, we randomly divide 16 samples into 8 uniform groups and obtain 8 prototypes by averaging features of 2 samples in each group. Considering such random division of samples might influence the performance, we experiment for 5 times and report the average scores.
The results from the third part of Table~\ref{alpha} illustrate that the more samples we cache, the higher accuracy Tip-Adapter can achieve.

\paragraph{Scaling Up to More Shots.} 
One might concern if more shots are given, the increasing size of the cache model along with higher-dimensional adapter would become a burden for both memory and computation. Thus, we here explore a way to fully utilize the training set with more than 16 shots, but fix the cache size as 16. Taking 64 shots as an example, following the division strategy in the above paragraph, we obtain 16 prototypes from 4 groups to construct the cache model. The results in the final part of Table~\ref{alpha} show that even if the cache size is restrained to 16, it still well contains the information of 32, 64 and 128 training samples per category. We also observe that the growth rate of performance gradually slows down when more samples are provided. This indicates a possible limit load for cache size 16, but Tip-Adapter-F solves it by fine-tuning and more shots result in larger performance boost.

\begin{figure}[t!]
\includegraphics[width=0.5\textwidth]{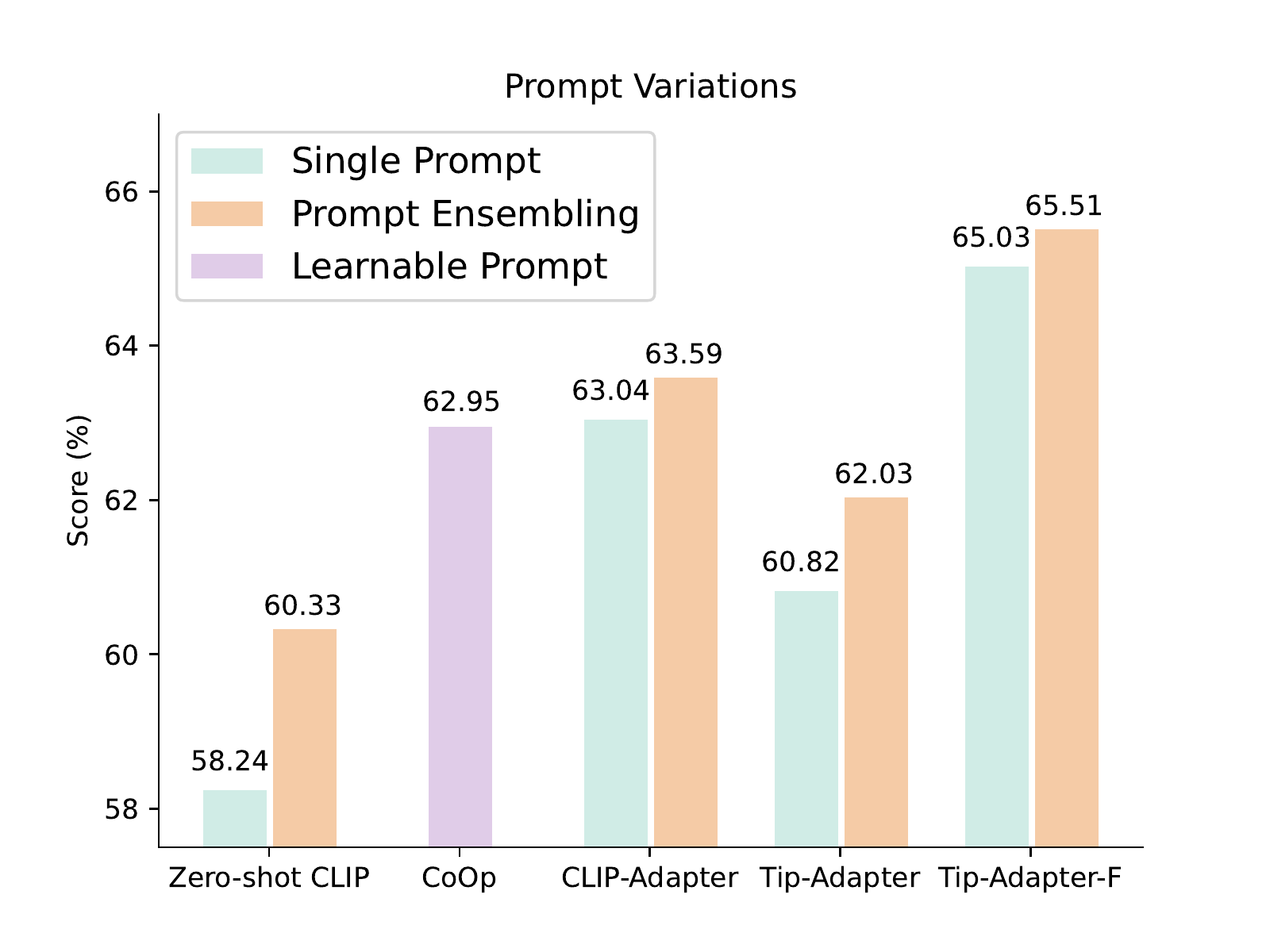}
   \caption{Performances of compared methods with three prompt designs: Single Prompt in cyan, Prompt Ensembling in orange and Learnable Prompt in purple.}
    \label{prompt}
    \vspace{-0.5cm}
\end{figure}

\paragraph{Prompt Design.}
For zero-shot CLIP, CLIP-Adapter and Tip-Adapter, the experiments above are all based on prompt ensembling of 7 templates from~\cite{radford2021learning}. In this study, we test only using a single prompt: ``a photo of a [CLASS].'' to see how this influences the performance. As shown in Figure~\ref{prompt}, the performance drops resulted from using a single prompt are smaller for Tip-Adapter-F and CLIP-Adapter, whose classification scores are relatively high, but larger for Tip-Adapter and Zero-shot CLIP. In general, the better-performing models are less affected by prompt variations.

\begin{table}[t!]
\centering
\vspace*{13pt}
\begin{adjustbox}{width=\linewidth}
	\begin{tabular}{cccccc}
	\toprule
		 Visual Enc. & Textual Enc. 
		 & Adapter $W_1$ & Adapter $W_2$ & Score($\%$)\\ \midrule
		- &- &- &- &62.03\\
		- &- &\Checkmark &- &65.51\\
		- & - & - & \Checkmark &60.90\\
		- & - & \Checkmark & \Checkmark &collapse\\
		\Checkmark & - & - & - &62.84\\
		- & \Checkmark & - & - & 63.15\\
	    \Checkmark & \Checkmark & - & - & 51.22\\
	\bottomrule
	\end{tabular}
\end{adjustbox}
\caption{Ablations of Tip-Adapter fine-tuning different modules. \Checkmark denotes fine-tuning the module and symbol - denotes freezing. Visual Enc. and Textual Enc. stand for visual encoder and textual encoder in pre-trained CLIP.}
\vspace*{-12pt}
\label{fine-tune}
\end{table}

\paragraph{Fine-tuning Settings.}
Tip-Adapter-F only fine-tunes adapter's $W_1$ in the cache model, but freezes $W_2$, CLIP's visual encoder and textual encoder. Here, we explore whether other modules in Tip-Adapter shall be fine-tuned. In Table~\ref{fine-tune}, we conduct 7 fine-tuning experiments on unfreezing different modules of Tip-Adapter. Note that we set the learning rate of the two CLIP encoders as 1/1000 of the learning rate of $W_1$. The first and second rows represent training-free Tip-Adapter's 62.03$\%$ and Tip-Adapter-F's 65.5$\%$. If we fine-tune $W_2$ in the adapter, the performance would fall to 60.90$\%$ or even collapse, which accords with our assumption that the one-hot ground-truth labels in $W_2$ shall not be updated. Furthermore, we fix all weights in the adapter and fine-tune the pre-trained CLIP's weights. If the visual encoder or textual encoder is independently tuned, the performance could be improved to 62.84$\%$ and 63.15$\%$, but when both encoders are jointly fine-tuned, the classification accuracy would significantly drop because of severe over-fitting, indicating that it is harmful to  fine-tune such a huge-parameter model with few-shot training set.

\subsection{Visualizations of Tip-Adapter}
In order to ease the understanding of cache-initialized Tip-Adapter and the variation during fine-tuning process, we utilize t-SNE ~\cite{radford2021learning} to visualize $W_1$, the first linear layer in the adapter or the cached training features. As shown in Figure~\ref{tsne}, dots in different colors denote 10 categories of ImageNet, and their relative distances here reflect the high-dimensional distributions of category embeddings. We conduct the visualization under the 16-shot settings, so there are 16 embeddings per category. 
From left to right in Figure~\ref{tsne}, the three sub-figures represent the initialized Tip-Adapter and the Tip-Adapter-F after fine-tuning, respectively. It could be observed that before training, such distribution has shown good discrimination thanks to the properly designed approach. During fine-tuning, embeddings of the same category gradually converges together and different clusters become more contrastive and separate with each other, contributing to stronger classification capability. This clustering process in feature space clearly demonstrates the performance improvement brought by Tip-Adapter-F.

\section{Conclusion}
\label{sec:conclusion}

We propose Tip-Adapter, the non-parametric extension of CLIP-Adapter, which obtains the weights of adapter not by SGD training but from a cache model constructed by few-shot training set. In this way, the feature extraction by adapter could also be viewed as retrieving the few-shot knowledge from the key-value cache. Surprisingly, Tip-Adapter with such training-free initialization achieves on par or even better results than CLIP-Adapter with training. In addition, Tip-Adapter could be further enhanced by fine-tuning for just a few epochs, a good trade-off between efficiency and performance. Considering limitations, although it is marginal, Tip-Adapter still requires 20-epoch fine-tuning to learn the best-performed weights. Our future work will focus on exploring new methods for adapter's weights construction to fully unleash its power for visual representation.

{\small
\bibliographystyle{ieee_fullname}
\bibliography{egbib}
}

\end{document}